# OPTIMIZATION OF CROSS DOMAIN SENTIMENT ANALYSIS USING SENTIWORDNET

K Paramesha[1] and K C Ravishankar[2]

[1]Department of Computer Science & Engg, Vidyavardhaka College of Engg., Mysore, India
paramesha.k@gmail.com
[2]Department of Computer Science & Engg, Government Engineering College, Hassan, India
kcrshankar@gmail.com

*ABSTRACT*

*The task of sentiment analysis of reviews is carried out using manually built / automatically generated lexicon resources of their own with which terms are matched with lexicon to compute the term count for positive and negative polarity. On the other hand the Sentiwordnet, which is quite different from other lexicon resources that gives scores (weights) of the positive and negative polarity for each word. The polarity of a word namely positive, negative and neutral have the score ranging between 0 to 1 indicates the strength/weight of the word with that sentiment orientation. In this paper, we show that using the Sentiwordnet, how we could enhance the performance of the classification at both sentence and document level.*

## 1. INTRODUCTION

The rapid strides made in the area of web technologies facilitated the customers to share their experiences of using the products (consumer products, books, movies…) across the globe. Such products could be purchased from vendors by the customers via online. These customers being the brand ambassadors of the products share their opinions (customer reviews) about product on suitable platform which will serve as feedback on products. The reviews are helpful to other prospective buyers who are planning to buy the products and the product manufacturing companies to make strategic decisions on products based upon the reviews. The present form of web provides multiple platforms such are blog spots, review forums etc. There are also different styles of rendering the feedback.

1) *Rating:* The overall product utility feedback is rate for a range (1-5) stars with 1 star being negative remark and 5 stars for positive remark. The rating sometimes made

2) *Thumbs up/down:* The positive feedback is indicated as thumbs up (+1) and negative for thumbs down (-1)

3) *Text Review:* The detailed elaborate user experience is scripted using natural language which is quite ambiguous in nature.

## 2. CROSS DOMAIN REVIEWS

The cross domain reviews refer to the collection of reviews on items, captured from different domains such as BOOK, DVDS, VIDEOGAMES ... Empirically it is observed that different words express sentiment in different domains. As mentioned in [2], same words may have different polarity in different domains. Difference in vocabularies across different domains also adds to the difficulty when applying classifiers trained on labeled data in one domain to test data in another. On employing a sentiment classifier, trained for a particular domain to classify sentiment of user reviews of a different domain often results in poor performance because words that occur in the train (source) domain might not appear in the test (target) domain [4]. In the *electronics* domain, we may use words like "compact", "sharp" to express our positive sentiment and use "blurry" to express our negative sentiment. While in the *video game* domain, words like "hooked", "realistic" indicate positive opinion and the word "boring" indicates negative opinion. In effect, with such of the words, the performance of the sentiment is greatly affected by the sentiment score.

## 3. DATA SET

The fine-grained sentiment dataset contains 294 product reviews from various online sources manually annotated with sentence level sentiment by Oscar and Ryan McDonald. The data is approximately balanced with respect to domain (books, DVDs, electronics, music, and videogames) and overall review sentiment (positive, negative, and neutral). In this work we considered only positive and negative reviews dataset. The dataset is available at [6]. The details of the review data set at document level and sentence level are furnished below in tabel1 and tabel2 respectively.

Table 1. Distribution of review documents

| Polarity | Domain | | | | | Total |
|---|---|---|---|---|---|---|
| | Books | DVDs | Electronics | Music | Video Games | |
| Positive Review | 19 | 19 | 19 | 20 | 20 | 97 |
| Negative Review | 29 | 20 | 19 | 20 | 20 | 99 |

Table 2. Distribution of review sentences

| Polarity | Domain | | | | | Total |
|---|---|---|---|---|---|---|
| | Books | DVDs | Electronics | Music | Video Games | |
| Positive Review | 160 | 164 | 161 | 183 | 255 | 923 |
| Negative Review | 195 | 264 | 240 | 176 | 442 | 1320 |

## 4. SENTIWORDNET

Sentiwordnet is a lexical resource for opinion mining. Sentiwordnet assigns to each synset of WordNet three sentiment scores: positivity, negativity, objectivity [1]. In this work we use SENTIWORDNET 3.0[3] which is an improved version of SENTIWORDNET 1.0, a lexical resource explicitly devised for supporting sentiment classification and opinion mining applications.

## 5. PREVIOUS WORK

Using the Sentiwordnet, [8] the common approach is based on term counting of both orientations. Sentiwordnet scores were calculated as positive and negative terms were found on each document and used to determine sentiment orientation by assigning the document to the class with the highest score. Further, using supervised learning methods, different aspects of text as sources of features have been proposed in the literature.

A total of 96 distinct features were generated based on the scores of positive and negative terms, for each parts-of-speech. The result of classification using feature set is slightly more than the term count result. The best result is just below 70%.

Applying the technique used by Brendan Tierney in [8], positive and negative scores for each review were calculated by counting positive and negative words, and then the sentiment polarity was determined by assigning the review to the class with the highest score. This method yielded an overall accuracy of 56.77%, with results detailed in the table

Sums on Review Positive and negative scores for each review were calculated using 'Sum on Review' technique as mentioned in [9] with various thresholds. This method yielded the results with the best accuracy of 67% at threshold 0.

Positive and negative scores for each review were calculated using Average on Sentence and Average on Review technique as mentioned in section 4 with various thresholds. This method yielded an overall accuracy of 68.63% at threshold 0. The results are detailed in table 3

## 6. OPTIMIZATION USING EVOLUTIONARY METHODS

For the optimization of classification, we consider an application of genetic algorithm (GA) which computes the weights of features. In prediction tasks, usually the datasets containing a large number of records and features that will be processed using, for example, created classification rules. Without pre-processing the dataset, i.e. without pre-weighting the attributes results in classification error. To counter the problem, GA is applied to find for each feature the weight that would reduce classification error value [7]. We perform the weighting of features with an evolutionary strategies approach using the variance of the Gaussian additive mutation adapted by a 1/5-rule. This rule [5] states that the ratio of successful mutations to all mutations should be 1/5, hence if the ration is greater than 1/5, the step size should be increased otherwise decreased.

$$x'_i = x_i + N(0, \sigma(t)) \qquad (1)$$

Where $N(0, \sigma(t))$ is a random Gaussian number with mean zero and mutation step $\sigma(t)$ is the function of $t^{th}$ generation. The simplest method is to specify the mutation mechanism is to use the same $\sigma(t)$ for all vectors in the population, for all variables for each vector. This will decrease slowly form weight 1 of parameters at the beginning of the run (t=0) to 0.1 as **t** approaches to the **T** generations. Using this mechanism, changes in the parameter values are based on the feedback from the fitness test and $\sigma(t)$ adaptation happens at every **n** generations. Such decreases may assist the fine tuning capabilities of the algorithm.

## 7. EXPERIMENTAL SETUP

The feature vector is generated using the Sentiwordnet 3.0 lexical resources as done in the [8]. The details of feature extract is tabled below. The sentences of review documents are POS tagged using Stanford Tagger utility to identify the POS of words. These tagged words are then produced to the Sentiwordnet 3.0 to generate positive and negative scores for the feature vector. The five different types of classifiers are applied for the sentiment analysis at document level and sentence level, with and without optimization of weights.

Table 3. Distribution of Feature Vector derived from SentiWordNet

| Category | Features | Count |
|---|---|---|
| Sentence Level | Sum of positive and negative scores and term count for Adjectives.<br><br>Sum of positive and negative scores and term count for Adverbs.<br><br>Sum of positive and negative scores and term count for Verbs.<br>Sum of positive and negative scores and term count for Nouns. | 16 |
| | Ratio of overall score per total terms found, for each part of speech. | 8 |

| | Sum of positive and negative scores and term count. | 4 |
|---|---|---|
| | Ratio of overall score per total terms found, for positive and negative words. | 2 |
| | Ratio of positive and negative scores.<br>Ratio of positive and negative terms. | 2 |
| Document Level | Summation of all features over sentences | 32 |

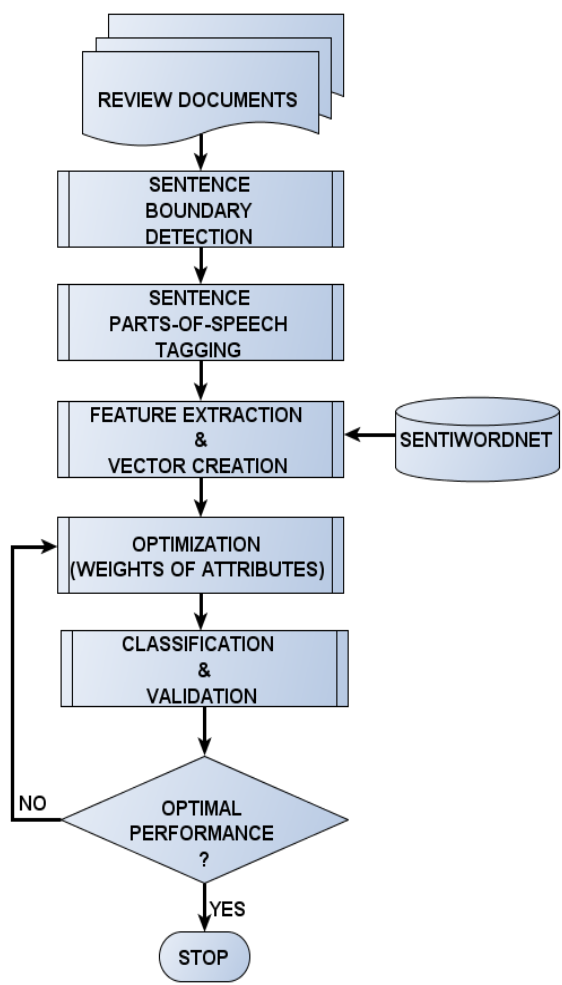

Figure 1. Steps Involved in Sentiment Analysis both at Sentence Level and Document Level

## 8. EXPERIMENT RESULT

Table 4. Experiment Results for Document Level Analysis (10-fold Cross Validation)

| classifier | Document Level | | | |
|---|---|---|---|---|
| | Without Optimization | | With Optimization | |
| | Accuracy | Precision | Accuracy | Precision |
| Svm | 65 | 73 | 72 | 81 |
| Naïve Bayes | 58 | 67 | 60 | 72 |
| Decision Tree | 43 | 38 | 52 | 67 |
| Linear Regression | 66 | 68 | 72 | 73 |
| Logistic Regression | 66 | 78 | 71 | 78 |

Table 5. Experiment Results for Sentence Level Analysis (10-fold Cross Validation)

| | Sentence Level | | | |
|---|---|---|---|---|
| classifier | Without Optimization | | With Optimization | |
| | Accuracy | Precision | Accuracy | Precision |
| SVM | 63 | 64 | 64 | 64 |
| Naïve Bayes | 63 | 56 | 54 | 58 |
| Decision Tree | 59 | 58 | 59 | 58 |
| Linear Regression | 65 | 62 | 66 | 63 |
| Logistic Regression | 65 | 67 | 65 | 68 |

## 9. DISCUSSION

On comparing the results obtained of this research and previous work results using Sentiwordnet, validates that Sentiwordnet could yield accuracy of 65%-75% with 10-fold cross validation, but from our observation, with single testing of whole weighted dataset and validation (without split validation) yields outstanding results which imply maximum convergence of machine learning algorithm. On other hand if the same optimized weighted data set is subjected to the 10-fold cross validation, the results were considerable lower due to overfitting and noise in the data set.

## 10. CONCLUSION

The experiment results shown implicate two important inferences, which are, validating the efficiency of the Sentiwordnet in discriminating the polarity of words both at sentence level and document level and second is the optimization techniques could lead to slightly better results for certain type of classifications. With the distinguishing ability of Sentiwordnet for cross domain words, in future work we would like to use features extracted from the Sentiwordnet as a compliment to other feature set derived from other techniques.

**Authors**

K.PARAMESHA
Assoc. Professor
Dept. of CSE, VVCE,
Mysore, India

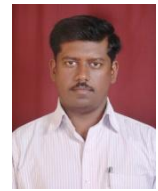

Studied BE in Computer Science (Bangalore university), M.Tech in Network Engineering (VTU, Belgaum). Have been working as a faculty in Dept. Of Computer Science at VVCE since 2000 and actively perusing PhD course in the area of Sentiment analysis.

Dr K.C.RAVISHANKAR
Professor & HOD
Dept. of CSE,
Govt. Engineering College
Hassan, India

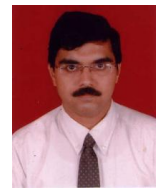

He has obtained BE in Computer Science and Engineering from Mysore University, M.Tech in Computer Science and Engineering from IIT, Delhi and doctorate from VTU, Belgaum in Computer Science and Engineering. He has 23 years of teaching experience. He served as Professor at Malnad College of Engineering Hassan.  Presently working as Professor & HOD at Govt. Engineering College Hassan, Karnataka. His area of interest include Image Processing, Information security, Computer Network and Databases.  He has presented a number of research papers at National and International conferences. Delivered invited talks and key note addresses at reputed institutions.  He has served as chairperson and reviewer for conferences and currently guiding three PhD scholars.